\documentclass[conference]{IEEEtran}
\IEEEoverridecommandlockouts
\usepackage{cite}
\usepackage{amsmath,amssymb,amsfonts}
\usepackage{algorithmic}
\usepackage{graphicx}
\usepackage{textcomp}
\usepackage{xcolor}

\usepackage{graphicx}
\usepackage{multirow}
\usepackage{amsmath,amssymb}
\usepackage{float}
\usepackage{xcolor} 
\usepackage{tcolorbox,xcolor} 

\usepackage{soul,xcolor} 
\usepackage[normalem]{ulem}
\setuldepth{+}

\usepackage{booktabs}
\usepackage{multirow}
\usepackage{tabularx}

\usepackage{soul}   

\setulcolor{red}
\usepackage{hyperref}

\def\BibTeX{{\rm B\kern-.05em{\sc i\kern-.025em b}\kern-.08em
    T\kern-.1667em\lower.7ex\hbox{E}\kern-.125emX}}
\begin{document}

\title{Multi-Objective Aligned Small Language Model Framework for SUD Patient Dialogue Generation\\
 }


\author{
\IEEEauthorblockN{Thushara Manjari Naduvilakandy, Hyeju Jang, Mohammad Al Hasan}
\IEEEauthorblockA{\textit{Department of Computer Science, Indiana University Indianapolis}\\
Indianapolis, IN, USA \quad
\{tnaduvil, hyejuj, alhasan\}@iu.edu}
}



\maketitle

\begin{abstract}
Substance Use Disorder (SUD) counseling requires patient responses that reflect
underlying cognitive states such as beliefs, coping strategies, and readiness
for change. Although large language models (LLMs) can generate fluent text, they
often fail to produce cognitively coherent and clinically realistic patient
behavior, especially under ethical and data-scarce clinical settings. Moreover, deploying frontier-scale LLMs in healthcare applications presents practical challenges including high computational cost, latency, privacy concerns, and limited deployability in resource-constrained environments, motivating the need for cognitively aligned small language models (SLMs). We propose
a cognitively grounded framework for SUD patient dialogue generation that
explicitly models and aligns latent cognitive components with patient histories
and counselor questions. Our pipeline consists of two stages: cognitive
component detection and cognitive component–aligned dialogue generation. To
enable effective learning with smaller models, we combine knowledge distillation
from high-capacity teacher models, preference optimization from
human-annotations, and attention-guided reward shaping. Extensive evaluations
using automatic scores like BERTScore, ROUGE, METEOR, and BLEU,  and LLM-as-judge
hit-metrics against both human and teacher-model references show that
cognitively informed fine-tuning substantially improves cognitive realization
and alignment over a generic instruction-tuned baselines and mental health domain specific SLMs, with particularly
strong gains for open-ended cognitive components. 
Github: \url{https://github.com/thusharamanjari/Multi_objective_aligned_SLM}

\end{abstract}


\section{Introduction}

Substance Use Disorder (SUD) is a leading cause of suicide, relapse, and
long-term mental health decline.
In the United
States, SUD has become a growing burden on public health with more than 932,000
deaths over the last two decades ~\cite{cdc2022overdose}. While behavioral
therapies like motivational interviewing (MI) and Cognitive Behavioral Therapy
(CBT)
are effective to overcome SUD, access to these
services is limited due to cost and counselor shortages~\cite{wang2005twelve},
specifically in rural areas where the SUD problem is particularly prevalent.
Often, patients are reluctant to take such services due to social stigma
associated with drug abuse~\cite{Mak2017StigmaDelay}.

AI-based systems can be a potentially accessible and stigma-free option for SUD
patients who are reluctant or unable to engage with human counselors for seeking
support. However, these patients are highly vulnerable, so an AI-driven chatbot,
must strive to provide safe and empathetically appropriate counseling.
Contemporary AI-driven chatbots are typically built using large language model platforms, 
but the applications of such platforms in healthcare settings have shown mixed
success; specifically, in mental health counseling, they exhibit shortcomings 
in clinical safety and empathetic understanding, with documented risks of
inappropriate or unsafe responses in high-stakes
interactions~\cite{sorin2024large}. 
These limitations
underscore the need for targeted alignment and fine-tuning to strengthen the ethical use of AI
 in mental health applications~\cite{shah2025evaluating}.

Building AI-driven chatbots for substance use disorder (SUD) counseling requires realistic AI patient simulators to safely train novice counselors amidst persistent staff shortages and strict privacy constraints. The objective of this work is to leverage LLMs to generate realistic SUD patient dialogues that reflect diverse backgrounds, histories, and cognitive structures. However, unlike generic open-domain conversations, SUD patient utterances are grounded in clinically meaningful, evolving psychological states—such as maladaptive beliefs, coping strategies, and readiness for change~\cite{beck2020cognitive}—that are heavily conditioned on the counselor’s therapeutic intent. While standard LLMs generate fluent text, they often fail to maintain consistency across these latent cognitive components, resulting in clinically incoherent behavior. Compounded by restricted access to real clinical data, this challenge necessitates the explicit modeling of psychological variance over reliance on generic conversational patterns~\cite{xu2022survey}.

In the current AI landscape, Small Language Models (SLMs) are becoming increasingly important due to the practical limitations associated with Large Language Models (LLMs), including high fine-tuning costs, substantial inference latency, privacy concerns, and substantial reliance on cloud infrastructure. In contrast, SLMs offer significant advantages in computational efficiency, reduced latency, lower deployment cost, and suitability for edge and on-device applications.
These properties are particularly relevant to SUD dialogue generation, which involves sensitive patient information and repeated interactions, making privacy, low-latency inference, and deployment cost important considerations. Recent studies further demonstrate that fine-tuned SLMs can achieve competitive—and in some domain-specific scenarios, even superior—performance compared to larger models, particularly when enhanced through domain adaptation, hybrid architectures, or targeted augmentation techniques. Moreover, fine-tuning large-scale LLMs remains computationally expensive and memory-intensive, limiting reproducibility and iterative experimentation in multi-stage training pipelines. Prior work shows that smaller models trained via knowledge distillation can retain much of the representational and reasoning capacity of larger models while being substantially more efficient; for instance, DistilBERT preserves approximately 95\% of BERT’s performance while using 40\% fewer parameters and enabling faster inference \cite{sanh2019distilbert}. More broadly, research suggests that full-scale LLMs are often unnecessary for many domain-specific generation tasks, with distilled or fine-tuned SLMs offering a more favorable balance between performance, controllability, and deployability \cite{hinton2015distilling}. 
Consequently, emerging evidence indicates that the most effective AI systems are not purely LLM-based, but instead integrate the efficiency of SLMs with selective LLM assistance to achieve improved performance, cost-effectiveness, trustworthiness, and scalability \cite{wang2025survey}.

We identify four key desiderata for an effective SLM-based SUD patient simulator: (i) automatically identifying cognitive components from patient history; (ii) explicitly modeling cognitive components of diverse patient profiles; (iii) aligning generated patient responses with counselor questions, underlying cognitive components, and dialogue length; and (iv) operating effectively under domain-specific data scarcity using smaller, non–frontier language models. Subsequently, we implement a cognitively grounded SLM-based pipeline that satisfies all four desiderata. Specifically, our simulator leverages cognitively rich patient histories to generate responses aligned with both counselor questions and patient cognitive components. By explicitly incorporating cognitive-component identification and alignment into the generation process, our approach enables psychologically coherent patient responses that remain consistent with patient history and therapeutic context without relying on large proprietary models. Methodologically, we show that supervision from high-capacity teacher models, human-annotated cognitive labels, and attention-guided reward-based optimization \emph{jointly} contribute to explicit, controllable cognitive alignment in smaller language models—both for cognitive component detection and downstream dialogue generation. To this end, we claim three research contributions spanning the design, development, and evaluation of our model.

Our first contribution is designing a two-stages framework for the
patient dialogue generation; the stages are: cognitive component detection and
cognitive component–aligned dialogue generation; in the first stage, we utilize
cognitively informed knowledge distillation and preference optimization for
cognitive component generation, specifically we examine whether fine-tuning with
teacher-model distillation and human-labeled preference supervision improve both
correctness with respect to human annotations and alignment with the teacher
model’s cognitive representations. In the second stage, we utilize supervised
fine-tuning, preference optimization, and reward shaping to enable smaller language models to generate dialogues that explicitly realize target cognitive components more effectively than generic instruction-tuned and domain-specific SLM baselines. 

Our second research contribution is to  examine the role of human-labeled
supervision for open-ended cognition, focusing on whether human-annotated
supervision for non-categorical cognitive components improves the realization of
open-ended cognitive structures derived from patient histories relative to
categorical label-based supervision. Finally, we explore whether reward-guided
alignment can encourage coherent grounding between generated patient dialogues,
cognitive-component spans, and counselor questions without inducing
lexical copying or reducing conversational naturalness.

Our third research contribution is extensive evaluation across both the stages
of our pipeline using automatic metrics (e.g., BERTScore and ROUGE),
LLM-as-judge analyses, and component-wise ablations against both human and
high-capacity LLM model references. Our results demonstrate that cognitively
informed fine-tuning substantially improves explicit cognitive realization and
controllable dialogue generation under constrained model capacity, while also
revealing clear performance headroom relative to teacher models.


\section{Related  works}

We break down the related works into two groups as discussed in the following subsection.

\subsection{AI Patient Works in NLG}

Recent work has used LLMs for AI patient simulation, therapist training, and controlled dialogue generation. PATIENT-$\Psi$~\cite{wang2024patient} uses manually extracted cognitive concepts (CCs) from patient histories to prompt GPT-4, while Roleplay-doh~\cite{louie2024roleplay} uses expert-defined behavioral principles to guide patient role-play. These approaches improve conversational fidelity and behavioral consistency but rely on static, human-defined concepts or rules and do not dynamically model latent patient cognitive states across dialogue turns.

Related NLG methods improve controllability through instruction- or strategy-aligned generation, such as Motivational Interviewing and predefined behavioral guidelines~\cite{gupta2022dialguide,sun2024rethinking}. Post-training methods further improve instruction following through SFT and preference optimization~\cite{ouyang2022training}. However, existing approaches largely control dialogue behavior or therapeutic strategy rather than explicitly learning and enforcing patient cognitive structure. 
More recently, ChatThero~\cite{wang2025chatthero} proposed a multi-agent LLM framework for addiction recovery support that integrates motivational interviewing and CBT strategies through simulated patient profiles and preference optimization. While ChatThero advances therapeutic dialogue generation and evaluation in the SUD domain, its primary focus is on end-to-end counselor–patient interaction rather than isolating and controlling patient utterances.



\begin{figure*}[ht]
 \small   
\setlength{\abovecaptionskip}{0.1em} 
    \centering
    \includegraphics[width=5in]{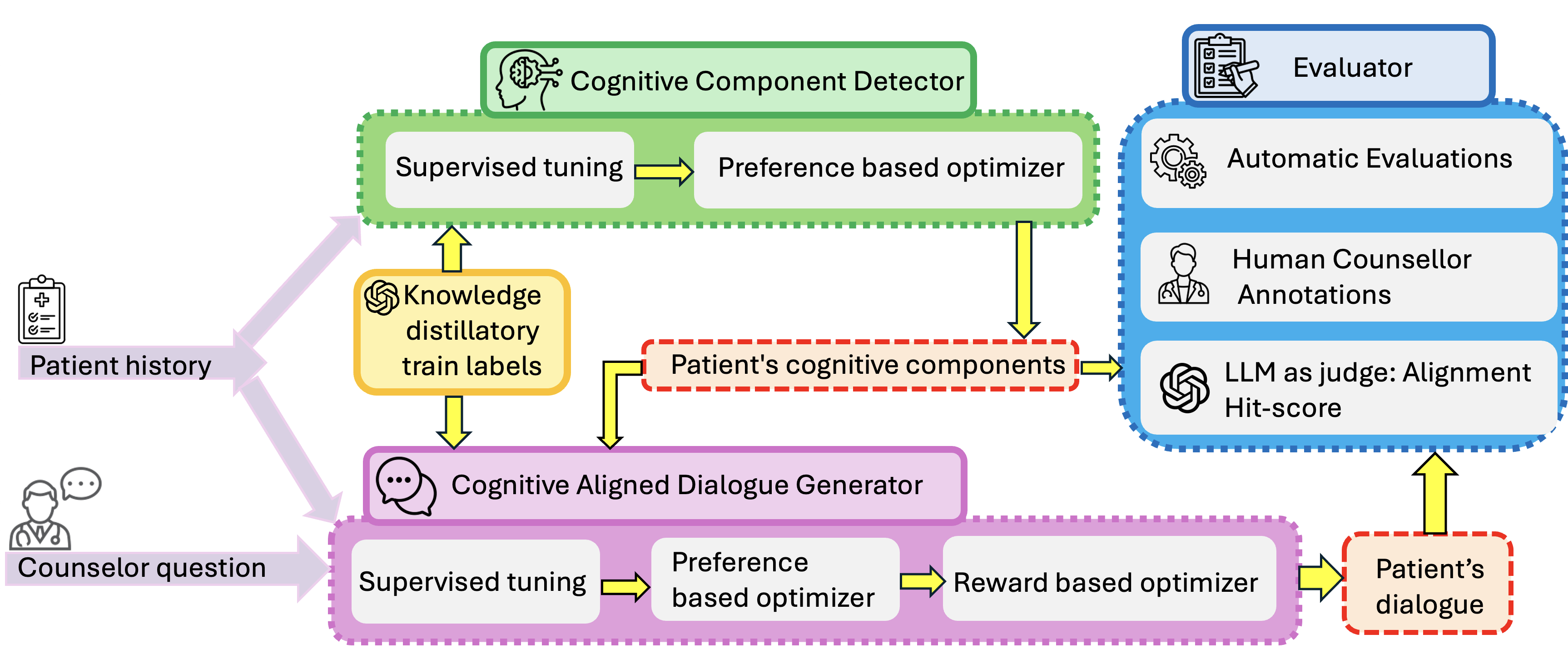}
    \caption{Framework for Cognitive Components-Aligned SUD Patient Dialogue
Generation in LLMs}
    \vspace{-0.2in}
    \label{fig:framework}
\end{figure*}

\subsection{SLMs for Conversational Agents}
Recent studies demonstrate the effectiveness of SLMs as backbones for conversational agents. Jun et al.~\cite{jun2025tod} show that SLMs distilled from large teacher models achieve competitive performance in task-oriented dialogue by learning from synthetic LLM-generated data, while Zheng et al.~\cite{zheng2024augesc} present AUGESC, an SLM-based emotional support chatbot that captures nuanced conversational strategies through distillation. Beyond text-only dialogue, Zhu et al.~\cite{zhu2024llavaphi} demonstrate that compact language backbones can support rich multimodal conversational assistants. Complementary analyses argue that SLMs are particularly well-suited for agentic and dialogue-centric applications, where efficiency and task specialization outweigh raw model scale~\cite{small2025agentic}, a conclusion reinforced by recent surveys and data-centric training efforts such as SmolLM2~\cite{smollm2025}
. Together, these findings motivate our use of an SLM as a practical and effective alternative to large proprietary LLMs for cognitively aligned dialogue generation.

\section{Methodology}

In Figure~\ref{fig:framework}, we present our end-to-end framework for Cognitive
Component–Aligned SUD patient dialogue generation. The method consists of two
core modules. The Cognitive Component Detector (CCD) (green module in the figure)
uses a smaller language model (SLM) trained via knowledge distillation,
supervised fine-tuning, and preference optimization to extract patient cognitive
components from patient histories. The Cognitive Component–Aligned Dialogue
Generator (CADG) (pink module) then generates patient responses conditioned on
the patient history, counselor question, and inferred cognitive components using
knowledge distillation, supervised fine-tuning, preference optimization, and 
attention-guided reward optimization. Following dialogue generation, a patient-specific style anchor module personalizes the responses with the patient's potential
linguistic style. At inference time, the trained framework takes a patient
history and a counselor question as input, identifies relevant cognitive components and
generates cognitively aligned patient responses. Lastly, an evaluation module (blue
module) assesses both the predicted cognitive components and the generated
dialogues using automatic metrics, human counselor annotations, and LLM-as-judge
alignment scores.

To construct information-rich synthetic patient histories for LLM alignment experiment, we leverage publicly available Reddit posts describing real-life substance-use experiences. Compared to randomly generated synthetic profiles produced solely via LLM prompting, such narratives provide substantially richer real-world context and a set of wide variety of Cognitive features. Details of patient history generation and statistics is provided
in Section~\ref{sec:dataset}.
In the subsequent sections we describe each of our modules in our framework.

\subsection{Cognitive Component Detector}

Cognitive Component Detector (CCD) module's main task is to generate cognitive
components from the patient history. It follows teacher training framework in
Large Language Models (LLMs), where a large, pre-trained "teacher" model guides
a smaller, more efficient "student" model (SLM) to learn faster, improve
accuracy, and distill complex reasoning skills. In our framework we use both
supervised fine-tuning and preference-based fine-tuning (a lightweight
alternative to reinforcement learning with human feedback) in sequential steps
to enable SLM to achieve superior performance on cognitive component detection.
In Figure ~\ref{fig:framework}, this component is shown in green color.

In collaboration with domain expert (licensed counselors) and following
professional practice as suggested in Beck\cite{beck2020cognitive}, we
identify 17 cognitive components (CCs)--psychological constructs critical for
modeling AI-simulated SUD patient behavior. These CCs are core belief,
intermediate belief, coping strategy, situation, automatic thought, emotion,
behavior, SUD stage, primary core belief, social issue, legal issue, financial
issue, drugs used, other mental health issues, physical health issues, past
trauma and DSM-5 traits. Using GPT-5 (teacher model) prompting we extract
these cognitive components from each patient history. We consider
these cognitive components as ground truth CC labels for knowledge distillation
and represent them as yellow module in Figure \ref{fig:framework}. 

\noindent\textbf{Supervised Fine-Tuning:} We first apply supervised fine-tuning
(SFT) to help the SLM (we used LLaMA-3.1-8B-Instruct model) learn explicit
input–output patterns from responses produced by teacher model
GPT-5 through knowledge distillation. This provides task-specific supervision when expert-annotated clinical data are scarce, following prior work on synthetic supervision in clinical NLP \cite{corbeil2025modular}. This initialization provides a stable
foundation for subsequent preference optimization. To enable the model to detect
cognitive components (CCs) from patient histories, we first design a set of 17
static instructions, one corresponding to each CC category. Using the 359 maximally informative patient
histories selected in previous step, we construct a supervised fine-tuning (SFT)
dataset. Each training instance consisted of: (i) an instruction specifying the
CC to be extracted, (ii) the patient history as input, and (iii) the ground
truth CC label(s) as output. In total, this produces 6,103 instances
(359 patients), split into  4,998 training (294 patient), 612 validation (36
patient), and 493 test samples (29 patient).

\noindent\textbf{Preference optimization:}
The motivation of using DPO (Direct Preference Optimization \cite{rafailov2023direct}) is to focus on
optimizing the model’s detection capability among cognitive components where
explicit domain cues are not provided in the instruction text (open ended
generations). 
For categorical cognitive components such as SUD stage, emotion, or core belief, the model
is provided with a discrete set of options drawn from established clinical
literature, allowing it to select the most appropriate label. This design
reflects an easier reasoning setup—comparable to how a non-expert human can make
accurate choices when guided by picking most suitable one from the predefined
categories. In contrast, cognitively generative features such as intermediate
belief, coping strategy, situation, automatic thought, and behavior require
deeper reasoning and implicit understanding of patient context. These components
cannot be accurately identified without interpreting the psychological dynamics
expressed in the patient history, which poses a challenge for both generic LLMs
and untrained annotators. 
To capture these subtleties, we construct preference
pairs where each of the five generative components is contrasted with the
remaining four, producing 20 pairwise preference instances per patient and a
total of 2,120 examples for DPO training. 

\vspace{-0.05in}
\subsection{Cognitive-Aligned Dialogue Generation}

The CCD module detects the cognitive components (CCs) of an unseen patient from their history. Based on these detected components, the Cognitive-Aligned Dialogue Generation (CADG) module (the pink module in Figure \ref{fig:framework}) generates aligned patient utterances. To train and validate this capability, we curated a fixed set of 17 counselor questions, each aligned with a specific CC. While real-world counseling is an iterative, multi-turn role-play rather than a fixed questionnaire, our experiments operate in a controlled, single-turn setting. This allows us to validate whether a simulated SUD patient can generate dialogue aligned with their history. Like the CCD module, CADG is trained using supervised fine-tuning and direct preference optimization. However, we also propose a novel reinforcement learning-based training paradigm through reward-based optimization. Each of these three training phases is discussed below.

\noindent\textbf{Supervised Fine-Tuning: }

We perform Supervised Fine-Tuning (SFT) to guide the SLM in producing instruction-following dialogues that emulate the explicit response patterns observed in the GPT-5-generated ground-truth data. Specifically, given a patient history, a cognitive component name, and the corresponding cognitive component value for the patient, the instruction requires the model to generate a patient dialogue aligned with the specified cognitive component. 
In this way, the SLM learns the structural and linguistic patterns by which GPT-5 realizes cognitively grounded patient responses. The training inputs consist of the patient history, cognitive component, and counselor question, while the output is the ground-truth dialogue from GPT-5. Since the output is natural language text, next token prediction loss is used for training.

\noindent\textbf{Preference Optimization: } Building on the SFT model, we apply Direct Preference Optimization (DPO) to refine the model’s ability to generate cognitively aligned dialogues. Similar to our Cognitive Component Detector module, this stage targets implicit cognitive constructs—such as \textit{intermediate beliefs, coping strategies, situational context, automatic thoughts, and behaviors}—which require deeper reasoning and an implicit understanding of patient context. Distinguishing between these categories is challenging without domain expertise in psychology, and thus purely supervised signals are insufficient.
To address this, we construct a preference dataset where, for each of the five implicit cognitive constructs, we pair one example against the remaining four, yielding 20 contrastive preference pairs per original instance. In total, we obtained 740 preference instances as input for DPO training.

\noindent\textbf{Reward-Based Optimization: } Reward-based optimization is particularly valuable in low-data settings where explicit supervision is limited but the desired behavioral outcomes are well defined. In our case, the objective is to train the model to generate patient utterances that are (i) aligned with the implicit cognitive component, (ii) responsive to the counselor’s question, and (iii) coherent with the patient’s history. Purely instruction-based fine-tuning on these three components often leads the SLM to overgeneralize or lose focus, as it lacks explicit feedback emphasizing the underlying cognitive intent of the dialogue.

To address this, we incorporate Group Reward Policy Optimization (GRPO)\cite{shao2024deepseekmath} stage that introduces a domain-specific reward function guiding the model toward cognitively aligned and contextually relevant responses. Specifically, we design an attention-guided reward mechanism that encourages the model to allocate higher attention to the cognitive component and counselor question segments of the input in a weighted manner. This helps the model maintain focus on the implicit cognitive components extracted by the CCD module. 
We initialize GRPO training from the DPO-aligned checkpoint, updating only the LoRA parameters while keeping the base model frozen. This stage reinforces cognitively grounded alignment through continuous reward feedback. 

For patient dialogue generation, CADG, module's input
comprises \textit{task description}, \textit{patient history},
\textit{cognitive feature}, and \textit{counselor question}. When processing
such long instruction context with multi-component input, the SLM struggles to
determine \textit{where} to focus its attention. 

To address this limitation, we propose an \textit{attention-guided reward} that quantifies how effectively the model ``reads from'' clinically salient sections of the prompt—namely, the task description ($\mathcal{T}$), patient history ($\mathcal{C}$), cognitive feature ($\mathcal{F}$), and counselor question ($\mathcal{Q}$). 
This reward leverages the model’s own self-attention patterns as a differentiable signal, encouraging generation that anchors in $\mathcal{F}$ and $\mathcal{Q}$ rather than diffusing attention across irrelevant regions. 

Let the model's last-layer attention tensor be 
$A^{(L)} \in \mathbb{R}^{H \times T \times T}$. 
Averaging across attention heads gives:
$
\bar A = \tfrac{1}{H}\sum_{h=1}^{H} A^{(L)}_h \in \mathbb{R}^{T\times T}.
$ In the attention matrix, each row corresponds to a \emph{query} token and each column to a \emph{key} token. 
We compute the attention from the generated tokens (queries) to the input tokens (keys). 
For each input token $j$, we average the corresponding attention values over all generated tokens $i \in Q_g$:
$
a_j = \tfrac{1}{|Q_g|}\sum_{i\in Q_g} \bar A_{ij}, \qquad j \in \{1,\ldots,T\}.
$ Thus $a_j$ quantifies how strongly the generated output reads from token $j$. We restrict the analysis to the set of all four prompt sections 
$U = \{ \mathcal{T}, \mathcal{C},  \mathcal{F}, \mathcal{Q}\}$ 
and renormalize the attention weights over these four section by grouping the token weights
of each section.
If $s_k$ denotes the normalized attention mass on section~$S_k \in U$, we define 
$
d_k=\frac{s_k}{|S_k|},$
where a higher $d_k$ indicates more concentrated attention per token.

Finally, when evaluating focus on the feature or question sections, we compute the mean densities of the other three sections as
$
\bar d_{\neg\mathcal{F}}=\tfrac{1}{3}(d_\mathcal{T}+d_\mathcal{C}+d_\mathcal{Q}),\qquad
\bar d_{\neg\mathcal{Q}}=\tfrac{1}{3}(d_\mathcal{T}+d_\mathcal{C}+d_\mathcal{F}).$ The attention-based reward $R_{\mathrm{attn}}$ (Eq.~\ref{eq:attn_reward}) has three components: base mass, density bonus, and geometric-mean synergy, with parameters $w_f{=}0.6$, $w_q{=}0.4$, $b{=}0.8$, $c{=}0.6$, and $K{=}0.9$. The \textit{base-mass} term rewards total attention on the cognitive-feature ($\mathcal{F}$) and counselor-question ($\mathcal{Q}$) sections. The weights $w_f{=}0.6$ and $w_q{=}0.4$ define their relative priority. The \textit{density-bonus} term
adds a reward when attention over $\mathcal{F}$ or $\mathcal{Q}$ is denser than that of other sections, encouraging focused attention. The coefficient $b{=}0.8$ controls this effect. The \textit{geometric-mean} term $c\,(s_\mathcal{F}^{\,w_f}s_\mathcal{Q}^{\,w_q})$ rewards simultaneous attention to both $\mathcal{F}$ and $\mathcal{Q}$, discouraging single-span focus. The coefficient $c{=}0.6$ sets the synergy strength, and $K{=}0.9$ scales the overall reward.

\begin{equation}
\vspace{-.5mm}
\begin{aligned}
R_{\mathrm{attn}}
&= K \Big[
\underbrace{w_f s_\mathcal{F} + w_q s_\mathcal{Q}}_{\text{base mass}}
+ \underbrace{c\,\big(s_\mathcal{F}^{\,w_f}
s_\mathcal{Q}^{\,w_q}\big)}_{\text{GM synergy}} \\
&\quad +
\underbrace{\tfrac{b}{2}\!\left(
\max\{0,d_\mathcal{F}-\bar d_{\neg\mathcal{F}}\}
+ \max\{0,d_\mathcal{Q}-\bar d_{\neg\mathcal{Q}}\}
\right)}_{\text{density bonuses}}
\Big].
\end{aligned}
\label{eq:attn_reward}
\end{equation}

\noindent\textbf{Length reward:} To maintain response coherence while avoiding excessively short or overlong completions, we introduce a \textit{length-based reward} $R_{\mathrm{len}}$, derived from the normalized word count $L$ of each generated completion after whitespace cleanup. The reward follows a piecewise linear design that penalizes brevity, rewards moderate verbosity, and saturates beyond a threshold:

\[
R_{\mathrm{len}} =
\begin{cases}
-1.0, & L \leq 30, \\[6pt]
\dfrac{L - 40}{40 - 30}, & 30 < L < 40, \\[10pt]
\dfrac{L - 40}{130 - 40}, & 40 \leq L \leq 130, \\[10pt]
1.0, & L > 130.
\end{cases}
\]

The computed reward is then scaled by a constant factor of $0.1$:
$
R_{\mathrm{len}}^{\mathrm{scaled}} = 0.1 \times R_{\mathrm{len}}.
$ This mild scaling ensures that the length reward contributes proportionally to the overall reinforcement signal—complementing higher-magnitude cognitive and attention-based rewards without dominating them. By rewarding moderate, information-rich responses and penalizing extreme brevity or verbosity, $R_{\mathrm{len}}$ effectively regularizes dialogue verbosity while preserving linguistic naturalness during GRPO optimization.

\section{Evaluation}

We evaluate our proposed framework at two stages of the pipeline: Cognitive Component Detection (CCD) and Cognitive Component–Aligned Dialogue Generation (CADG). 

\subsection{Datasets }
\label{sec:dataset}
To obtain a large corpus of realistic patient histories without relying on purely fabricated content, we scrape 1,828 posts from seven drug addiction-related subreddits using reddit's OAuth token and \texttt{Praw} Python package: \textit{r/DrugAddiction (500 posts)}, \textit{r/leaves}(500 posts), \textit{r/addiction} (500 posts), \textit{r/drugaddictionrecovery} (84 posts), \textit{r/drugaddicts} (96 posts), \textit{r/drug} (140 posts), and \textit{r/marijuanaaddiction} (8 posts). The collected posts are subsequently filtered using an LLM-based prompting procedure to retain only those containing sufficient personal or relational narratives suitable for reconstruction as patient histories. 
Each retained post is then transformed into a professionally written patient-history-style narrative using \textsc{Llama-3-8B}. 
The resulting corpus consists of 366 anonymized and standardized synthetic patient histories.



\noindent\textbf{Data split. }
We split the 366 patient histories into training, validation, and test sets containing 294, 36, and 36 patients, respectively. After the initial split, the test set is further manually filtered to ensure alignment with real-world patient self-reporting standards. Specifically, we remove patient histories that are not narrated in the first person by the patient, but instead written by friends or relatives, as such narratives do not reflect authentic patient cognitive framing. As a result, 7 test instances are excluded, yielding a final test set of 29 patient histories.


\subsection{Experimental Setup}

For supervised fine-tuning of both the Cognitive Component Detection (CCD) and
Cognitive Component--Aligned Dialogue Generation (CADG) modules, we employ
parameter-efficient LoRA adapters with 4-bit quantization. During training,
only the adapter weights are updated while the base model remains frozen. We
use a learning rate of $1 \times 10^{-4}$, a per-device batch size of 2,
gradient accumulation over 4 steps, the AdamW 8-bit optimizer, and a linear
learning rate scheduler for 3 epochs. All fine-tuning
experiments are conducted on an NVIDIA H100 80GB GPU. The best-performing
checkpoint from this supervised stage is saved and used for preference-based
optimization in the subsequent training stage.

\noindent\textbf{Evaluation Metrics.}
To assess the quality of the generated Cognitive Components (CC) and patient dialogues, we employ several standard NLP metrics. Lexical metrics, such as {\bf ROUGE-1/2/L, METEOR}, and {\bf BLEU} measure explicit surface-level overlap between the generated CC and reference CC.
We use them to assess whether the generated CCs are expressed explicitly using canonical labels or by salient phrases. These are particularly informative for identifying the ``explicit realization'' of cognitive constructs that generic LLMs often miss. Unlike exact string-matching, {\bf BERTScore} 
utilizes contextual embeddings from BERT to capture semantic similarity. This allows us to reward generated outputs that are meaningfully correct even if they use different wording than the reference labels.

Similarity-based metrics reward textual proximity but cannot reliably detect missing, paraphrased, or over-generated cognitive components. To address this limitation, we employ an LLM-as-a-judge evaluation framework using \textsc{GPT-5-mini}. The judge serves strictly as a binary verifier against fixed reference CCs (expert- or GPT-5-labeled). This judge assesses the alignment between a reference cognitive component set $C = \{c_1, c_2, \ldots, c_n\}$ and a generated cognitive component set $S = \{s_1, s_2, \ldots, s_m\}$ (for CC detection) or dialogue sentences (for generation).  
For each element in $C$, the judge determines if a semantically corresponding element exists in $S$. The reference CCs remain fixed throughout evaluation; the judge only performs binary classification and does not generate, modify, or select them. We then define the following ratios to quantify performance. First, {\bf Ref-Hit Ratio} measures coverage (recall) by calculating the proportion of reference components successfully reflected in the output:$\text{Ref-Hit Ratio} = \frac{|C_{\text{matched}}|}{|C|}$.



\begin{table*}[t]
\caption{Cognitive Component Detection (CCD) performance across BERTScore, ROUGE, METEOR, and BLEU.}
\label{tab:ccdi_rouge}
\centering
\footnotesize
\setlength{\tabcolsep}{2.5pt}
\renewcommand{\arraystretch}{0.95}

\begin{tabular}{@{}lcccccccccccc@{}}
\toprule
\multirow{2}{*}{\textbf{Generations}} 
& \multicolumn{12}{c}{\textbf{Reference Labels}} \\
\cmidrule(lr){2-13}
& \multicolumn{6}{c}{\textbf{Human}} 
& \multicolumn{6}{c}{\textbf{GPT-5}} \\
\cmidrule(lr){2-7} \cmidrule(lr){8-13}
& \textbf{BERT-F1} & \textbf{R-1} & \textbf{R-2} & \textbf{R-L} 
& \textbf{METEOR} & \textbf{BLEU}
& \textbf{BERT-F1} & \textbf{R-1} & \textbf{R-2} & \textbf{R-L} 
& \textbf{METEOR} & \textbf{BLEU} \\
\midrule

Llama3-8B-instruct   
& 0.805 & 0.0289 & 0.0074 & 0.0258 & 2.31 & 0.10
& 0.813 & 0.0666 & 0.0191 & 0.0515 & 6.81 & 0.35 \\

MentaLLaMA7B   
& 0.812 & 0.0259 & 0.0047 & 0.0221 & 1.90 & 0.09
& 0.819 & 0.0617 & 0.0150 & 0.0463 & 5.49 & 0.30 \\

MentaLLaMA13B   
& 0.815 & 0.0243 & 0.0037 & 0.0210 & 1.61 & 0.09
& 0.819 & 0.0576 & 0.0136 & 0.0450 & 4.84 & 0.35 \\

MentaLLaMA13B-Fewshot   
& 0.828 & 0.0678 & 0.0069 & 0.0647 & 3.46 & 0.78
& 0.830 & 0.1018 & 0.0185 & 0.0923 & 6.94 & 1.33 \\

\textbf{CCD}       
& \textbf{0.833} & \textbf{0.3046} & \textbf{0.0273} 
& \textbf{0.3007} & \textbf{12.00} & \textbf{3.80}
& \textbf{0.832} & \textbf{0.3687} & \textbf{0.0566} 
& \textbf{0.3470} & \textbf{17.37} & \textbf{5.06} \\

\midrule

GPT-5       
& 0.860 & 0.4896 & 0.0612 & 0.4840 & 19.51 & 6.69
& -- & -- & -- & -- & -- & -- \\

\bottomrule
\end{tabular}
\end{table*}

\begin{table}[t]
\caption{Cognitive Component Detection (CCD)  LLM-Judge Ref-Hit Ratio.}
\centering
\small
\setlength{\tabcolsep}{2.5pt}
\renewcommand{\arraystretch}{0.95}

\begin{tabular}{lc}
\toprule
\textbf{Gen vs Ref} 
& \textbf{Ref-Hit ratio (PM / FM / NM \%)} \\
\midrule

\textbf{CCD vs Expert}
& 48.3 / 45.4 / 51.7 \\

\textbf{CCD vs GPT-5}
& 60.0 / 57.4 / 40.0 \\

\textbf{GPT vs Expert}
& 61.5 / 57.8 / 38.5 \\

\bottomrule
\end{tabular}

\label{tab:CCDI_LLMjudgeScore}
\end{table}

\begin{table}[t]
\centering
\caption{Per–cognitive component difference in reference hit ratio between GPT-5 and CCD when evaluated against human annotations ($\Delta R = R_{\text{GPT}} - R_{\text{CCD}}$). Positive values indicate higher GPT coverage, while negative values indicate higher CCD coverage.}
\label{tab:deltaR_CCwise}
\small
\setlength{\tabcolsep}{2.5pt}
\renewcommand{\arraystretch}{1.1}
\begin{tabular}{lclc}
\toprule
\textbf{CC} & $\boldsymbol{\Delta R}$ 
& \textbf{CC} & $\boldsymbol{\Delta R}$ \\
\midrule
Intermediate belief        & $-0.1034$ 
& DSM-5 traits              & $0.1034$ \\

Legal issue                & $-0.0690$ 
& Physical health issues    & $0.1379$ \\

Automatic thought          & $-0.0690$ 
& Primary core belief       & $0.1552$ \\

Core belief                & $-0.0632$ 
& Social issue              & $0.1552$ \\

Emotion                    & $-0.0517$ 
& Drugs used                & $0.2011$ \\

Financial issue            & $-0.0345$ 
& SUD stage                 & $0.3103$ \\

Situation                  & $0.0000$ 
& OtherMentalHealthIssues   & $0.5345$ \\

Behavior                   & $0.1034$ 
& Past trauma               & $0.7241$ \\

Coping strategy            & $0.1034$ 
& \\
\bottomrule
\end{tabular}
\end{table}

\begin{table}[t]
\caption{Cognitive Component Aligned Dialogue Generation (CADG) BERTScores.}
\label{tab:CCDAG_bertscore}
\centering
\footnotesize
\setlength{\tabcolsep}{1.8pt}
\renewcommand{\arraystretch}{0.95}

\begin{tabular}{@{}lccc@{}}
\toprule
\textbf{Generations} 
& \textbf{F1} 
& \textbf{P} 
& \textbf{R} \\
\midrule

Llama3-8B-instruct   
& $0.829 \pm 0.013$
& $0.807 \pm 0.016$
& $0.853 \pm 0.014$ \\

MentalLlama7B   
& $\textbf{0.856} \pm \textbf{0.016}$
& $\textbf{0.851} \pm \textbf{0.020}$
& $0.862 \pm 0.020$ \\

MentalLlama13B
& $0.850 \pm 0.024$
& $0.838 \pm 0.034$
& $0.864 \pm 0.019$ \\

MentalLlama13B-Fewshot   
& $0.850 \pm 0.024$
& $0.840 \pm 0.034$
& $0.864 \pm 0.020$ \\

\textbf{CADG}       
& $0.844 \pm 0.018$
& $0.826 \pm 0.027$
& $\textbf{0.865} \pm \textbf{0.015}$ \\

\bottomrule
\end{tabular}
\end{table}
\vspace{-0.05in}

\begin{table*}[t]
\caption{Cognitive Component Aligned Dialogue Generation (CADG) performance across various reference labels and baseline models. PM / FM / NM (\%) represents Partial Match, Full Match, and No Match ratios respectively.}
\centering
\small
\footnotesize
\setlength{\tabcolsep}{2.5pt}
\renewcommand{\arraystretch}{0.95}

\begin{tabular}{lccc}
\toprule
\textbf{Generated Model} & \textbf{GPT-5 Reference(370)} & \textbf{Expert Reference(249)} & \textbf{CCD Reference(350)} \\
& PM / FM / NM (\%) & PM / FM / NM (\%) & PM / FM / NM (\%) \\
\midrule
\textbf{CADG (Ours)}    & \textbf{56.8}  /\textbf{55.4}  /\textbf{43.2 } & \textbf{49.3}/  \textbf{48.1} / \textbf{50.7} & 97.1 /80.9  /2.9  \\
Llama8b-instruct   &54.0/ 52.9  / 46.0& 45.4 / 44.4 /54.6 &  --  \\
MentaLlama-7B    & 47.1 / 46.2 /52.9  & 39.4/38.1 /60.6 &  --  \\
MentaLlama-13B    & 49.9 / 48.9 / 50.1 &  42.4 / 41.8/57.6&  --  \\
MentaLlama-13B+FS   & 52.7 / 51.9 / 47.3 & 46.2  /44.4/  53.8& --  \\

\bottomrule
\vspace{-0.2in}
\end{tabular}
\label{tab:updated_CADG_results}
\end{table*}

\vspace{-0.05in}

\begin{table}[t]
\caption{Automatic evaluation scores with ROUGE,BLEU and  METEOR for human reference cognitive alignment analysis. Showing our CADG got low rouge score than other baselines except Llama 8B instruct and still achieves high LLM judge alignment score indicating model did not copy paste the CC but it paraphrased for alignment.}
\centering
\small
\footnotesize
\setlength{\tabcolsep}{2.5pt}
\renewcommand{\arraystretch}{1.1}

\begin{tabular}{lcccc}
\toprule
\textbf{Model} 
& \textbf{ROUGE (R-1 / R-2 / R-L)} 
& \textbf{BLEU} 
& \textbf{METEOR} \\
\midrule
Llama8b-instruct        
&0.012/0.003  /0.010    
& 0.03 & 1.55  \\
Mentallama-7B        
&0.024  /0.004  /0.019  
&0.07  & 2.29 \\
Mentallama-13B        
& 0.025 / 0.005  / 0.020 
& 0.08 & 2.39 \\
Mentallama-13B+FS        
& 0.021 / 0.003  /  0.017
&0.06  & 2.02  \\

\textbf{CADG}                    
& 0.018 / 0.004 / 0.016 
&0.04  & 1.67 \\

\bottomrule
\vspace{-0.2in}
\end{tabular}

\label{tab:ccdag_rouge}
\end{table}



\begin{table*}[t]
\centering
\small
\footnotesize
\setlength{\tabcolsep}{2.5pt}
\renewcommand{\arraystretch}{1.1}

\caption{Analysis on GRPO reward ablation study on Cognitive Component attention Counselor Question attention and Length reward (reports average response length)}

\begin{tabular}{lccccccc}
\toprule
\multirow{2}{*}{\textbf{Model}} 
& \multicolumn{3}{c}{\textbf{Cognitive Component }} 
& \multicolumn{3}{c}{\textbf{Counselor Question }} 
& \textbf{Length  } \\
\cmidrule(lr){2-4} \cmidrule(lr){5-7} \cmidrule(lr){8-8}
& \textbf{ROUGE (R1/R2/RL)} & \textbf{BLEU} & \textbf{METEOR}
& \textbf{ROUGE (R1/R2/RL)} & \textbf{BLEU} & \textbf{METEOR}
& \textbf{\#words$<40$/\#words$>$40} \\
\midrule

\textbf{CADG}                    
& 0.056 / 0.034 / 0.050 & 1.05 & 6.75
& 0.093/0.030/0.068 &0.93 &14.63 
& \textbf{0.0\%} / \textbf{100.0}\% \\

CConlyAttendRwd 
& \textbf{0.532 / 0.485 / 0.533 }& \textbf{20.49} & \textbf{28.76}
& 0.028/0.003/0.025 &0.25 &1.36 
& 98.4\%/ 1.6\% \\

CQonlyAttendRwd 
& 0.023/0.001/0.020 & 0.15 & 1.15
& \textbf{0.647/0.554/0.611} & \textbf{33.45} &\textbf{48.40} 
& 94.7\% / 5.3\% \\

lengthOnlyRwd 
& 0.047/ 0.024/ 0.038 & 0.52 & 5.86
& 0.081/ 0.022/ 0.058 & 0.54 & 13.63
& \textbf{0.0\% / 100.0\%} \\

\bottomrule
\end{tabular}

\label{tab:grpo_ablation}
\end{table*}

\subsection{CCD Evaluation }

In Table~\ref{tab:ccdi_rouge}, we report ROUGE-1/2/L, METEOR, and BLEU scores
measuring lexical and phrase-level overlap between generated cognitive component
(CC) text and reference labels. The reference labels were generated by GPT-5 and
human annotations; the latter were generated by the majority agreement among
three domain experts. 
In this table, CCD denotes the fine-tuned outputs from our Cognitive Component Detection module, which we compare against five systems: Llama3-8B-Instruct, MentaLLaMA7B, MentaLLaMA13B, MentaLLaMA13B-Fewshot, and GPT-5. All four base models—Llama3-8B-Instruct, MentaLLaMA7B, MentaLLaMA13B, and GPT-5—are evaluated under the same prompting setup to detect cognitive components from the test patient histories. MentaLLaMA13B-Fewshot additionally receives a dynamic one-shot example consisting of a patient history paired with its cognitive components, where the example is selected based on the targeted cognitive component of the current instance.

As can be seen from this table, the baselines exhibits
consistently low ROUGE, METEOR, and BLEU scores under both human and GPT-5
reference settings, indicating minimal explicit lexical realization of cognitive
components. This behavior is expected, as the baseline often conveys cognitive
information implicitly rather than producing exact cognitive phrases. In
contrast, the CCD fine-tuned model substantially improves all three metrics
across reference settings, indicating that cognitively informed supervised and
preference tuning encourages more explicit generations of cognitive expressions.
 
Under human annotations, CCD improves from the best performing baseline from a ROUGE-L score of 0.06 to 0.30, METEOR from 3.41 to 12.00, and BLEU from 0.78 to 3.80, with similarly strong gains observed when evaluated  against GPT-5 reference labels.
GPT-5 achieves the highest lexical and phrase-level overlap when evaluated against human annotations, serving as a high-capacity upper bound.

In Table~\ref{tab:ccdi_rouge}, we also report BERTScore F1 value
averaged over the entire dataset for cognitive component (CC) detection across different methods.  
The baseline models achieves moderately high BERTScores under both human and GPT-5 reference settings, indicating partial semantic alignment with reference cognitive components despite the absence of explicit cognitive fine-tuning. Our CCD model consistently improves F1 relative to the baseline across both reference settings, demonstrating that cognitively informed supervised and preference tuning enables the smaller 8B model to better realize and semantically align cognitive components with the reference labels.

In Table~\ref{tab:CCDI_LLMjudgeScore}, we report the values of Ref-Hit-Ratio, as
defined earlier. In this table, we compare CCD vs Expert (Human), CCD vs GPT-5, and GPT
vs Expert. Since one CC of a patient (one instance) may have multiple values, we
compute FM (full match), PM (partial match), and NM (no match) where PM and NM are mutually exclusive, while PM and FM may overlap. For example,
45.4\% of CCs/instance in our test set have fully matched between CCD and Expert, but 
this number improves to 57.4\% when compared CCD with GPT-5. The higher full match 
number with GPT-5 is due to the fact that our proposed CCD module is aligned with 
the GPT-5 teacher model, whereas it is not aligned with the Expert (human annotations).
The results in the third row, where GPT is compared against human annotation, the
full match is only 57.8\%, which shows that GPT's detection of human annotated CC, 
although better than CCD, it has much room for improvement. The trend for partial
match (PM) is similar to FM, and the trend for no match (NM) is obviously the 
opposite of the other two kind of matches. 



\subsubsection{CCD component importance: Role of SFT and DPO: }

Supervised fine-tuning (SFT) improves instruction following by encouraging the model to output only comma-separated cognitive components, without extraneous preambles. Compared to the baseline model, SFT substantially increases lexical alignment with the reference GPT outputs: the ROUGE-L score for open-ended outputs improves from 0.11 to 0.25, and the ROUGE-L score for closed-ended outputs improves from 0.02 to 0.62. These improvements indicate substantially stronger alignment with the GPT-annotated cognitive components and better adherence to the prescribed output format.

Preference optimization(DPO) helps to refine the quality of open-ended cognitive components. In Table~\ref{tab:deltaR_CCwise}, we report the per–cognitive component difference in reference hit ratio between CCD and GPT-5 when evaluated against human annotations. Negative $\Delta R$ values indicate components for which CCD achieves higher or comparable coverage, while positive values indicate components where GPT-5 achieves higher coverage. GPT-5 shows substantially higher hit ratios for components such as past trauma, other mental health issues, and SUD stage.
In contrast, CCD achieves comparable or higher coverage for  intermediate beliefs, automatic thoughts, situation, behavior and core beliefs. These components belong primarily to the open-ended generation category. And they have undergone both supervised fine-tuning and preference optimization during DPO, and consequently, components trained through this two-stage fine-tuning process either outperform GPT-5 or exhibit only marginal degradation relative to GPT-5. This trend indicates that preference optimization is effective for open-ended, cognitive components. In parallel BERTScore for open ended cognitive components increases from 0.8781 for the supervised tuned model to 0.8790 for the supervised tuning + preference optimization model. This both metrics results shows that preference optimization improved the quality of open-ended cognitive components.


\subsection{CADG Evaluation}
In Table~\ref{tab:CCDAG_bertscore}, we report mean and standard deviation of BERTScore computed against GPT-generated cognitive–aligned dialogues for dialogues generated by the baselines and our framework at the cognitive–aligned dialogue generation (CADG) module. All baselines Llama3-8B-Instruct, MentaLLaMA7B, and MentaLLaMA13B and our framework generate patient responses from the same inputs, namely the patient history and counselor questions. MentaLLaMA13B-Fewshot additionally receives a single example consisting of a patient history, counselor question, and GPT-generated dialogue; this example is selected dynamically by matching the cognitive component of the current instance.
Across the test set, CADG achieves a BERTScore F1 comparable to the best-performing domain-adapted baselines (e.g., MentaLLaMA), indicating that explicit cognitive fine-tuning can match the semantic quality of specialized mental-health models.

Table~\ref{tab:updated_CADG_results} reports Ref-Hit Ratio–based LLM-judge results comparing our CADG module against baseline models under two reference settings: GPT-5 cognitive component labels and expert-annotated cognitive component labels. Across both reference types, CADG attains the highest partial and full-match rates and the lowest no-match rates among all evaluated models, indicating consistently stronger coverage of the intended cognitive components than the domain-adapted MentaLLaMA baselines.

When evaluated against our own CCD-predicted cognitive labels, CADG achieves a Ref-Hit Ratio profile of 97.1\% partial match, 80.9\% full match, and only 2.9\% no match, reflecting near-complete recall of targeted components and serving as an approximate upper bound on performance under this metric. Overall, these results suggest that cognitively grounded dialogue generation substantially improves coverage of intended cognitive components, even under challenging reference conditions defined by GPT-5 and expert labels.

In Table~\ref{tab:ccdag_rouge}, we report ROUGE-1/2/L, BLEU, and METEOR scores between generated patient responses and the cognitive component (CC) text included in the prompt to guide the dialogue. These overlap-based metrics remain low across all models, reflecting the fact that cognitively aligned responses are expected to paraphrase rather than verbatim copy the CC text. Nevertheless, CADG attains ROUGE and METEOR scores that are comparable to or better than most baselines while simultaneously achieving strong LLM-judge alignment, indicating that the model preserves the intended cognitive content primarily through paraphrasing rather than direct reuse of the prompt CCs.

\begin{figure}[h]
 \small   
\setlength{\abovecaptionskip}{0.1em} 
    \centering
    \includegraphics[width=3in]{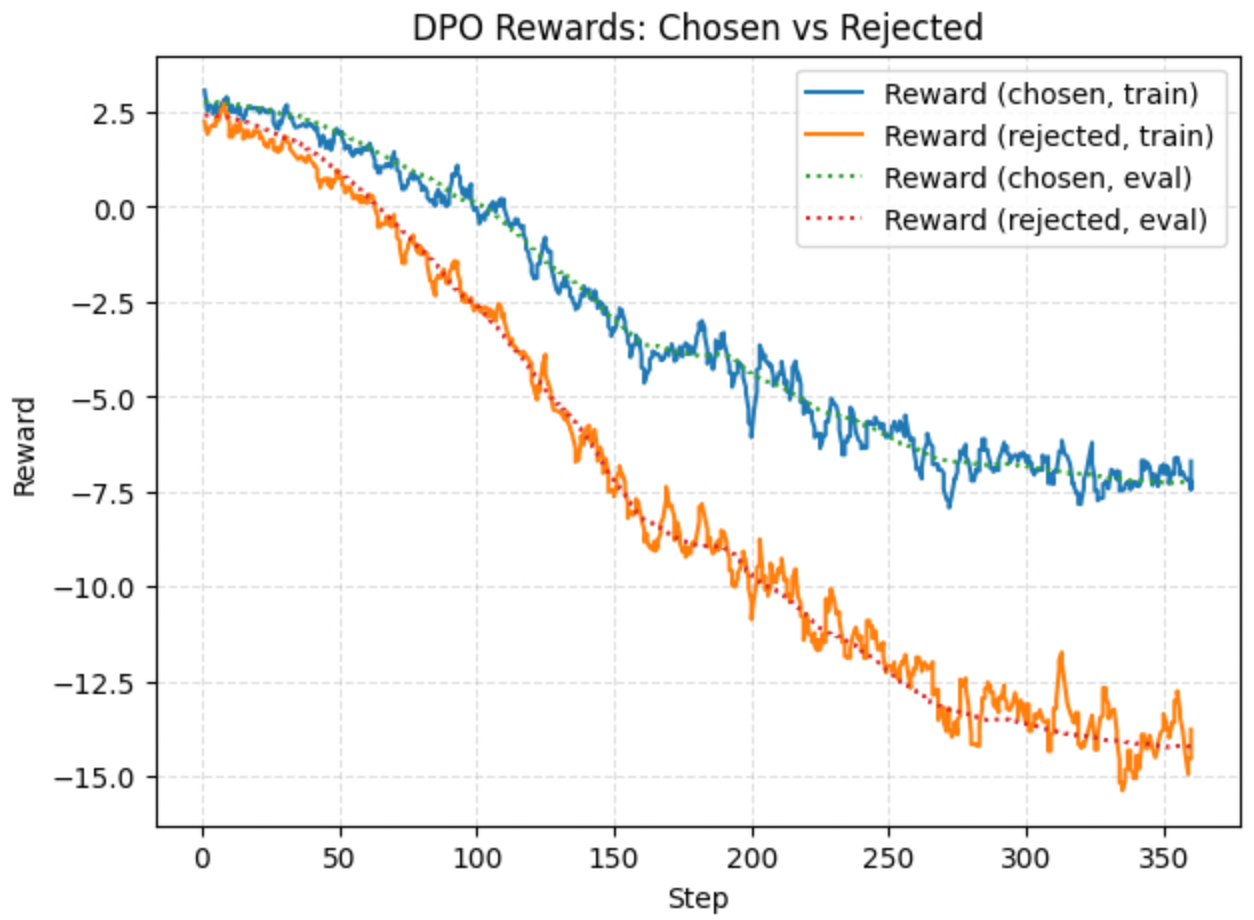}
    \caption{Training reward progress of preference optimization (DPO) for chosen and rejected in cognitive aligned dialogue generator.}
    \vspace{-0.2in}
    \label{fig:DPO_dial_reward}
\end{figure}

\begin{figure}[h]
 \small   
\setlength{\abovecaptionskip}{0.1em} 
    \centering
    \includegraphics[width=3in]{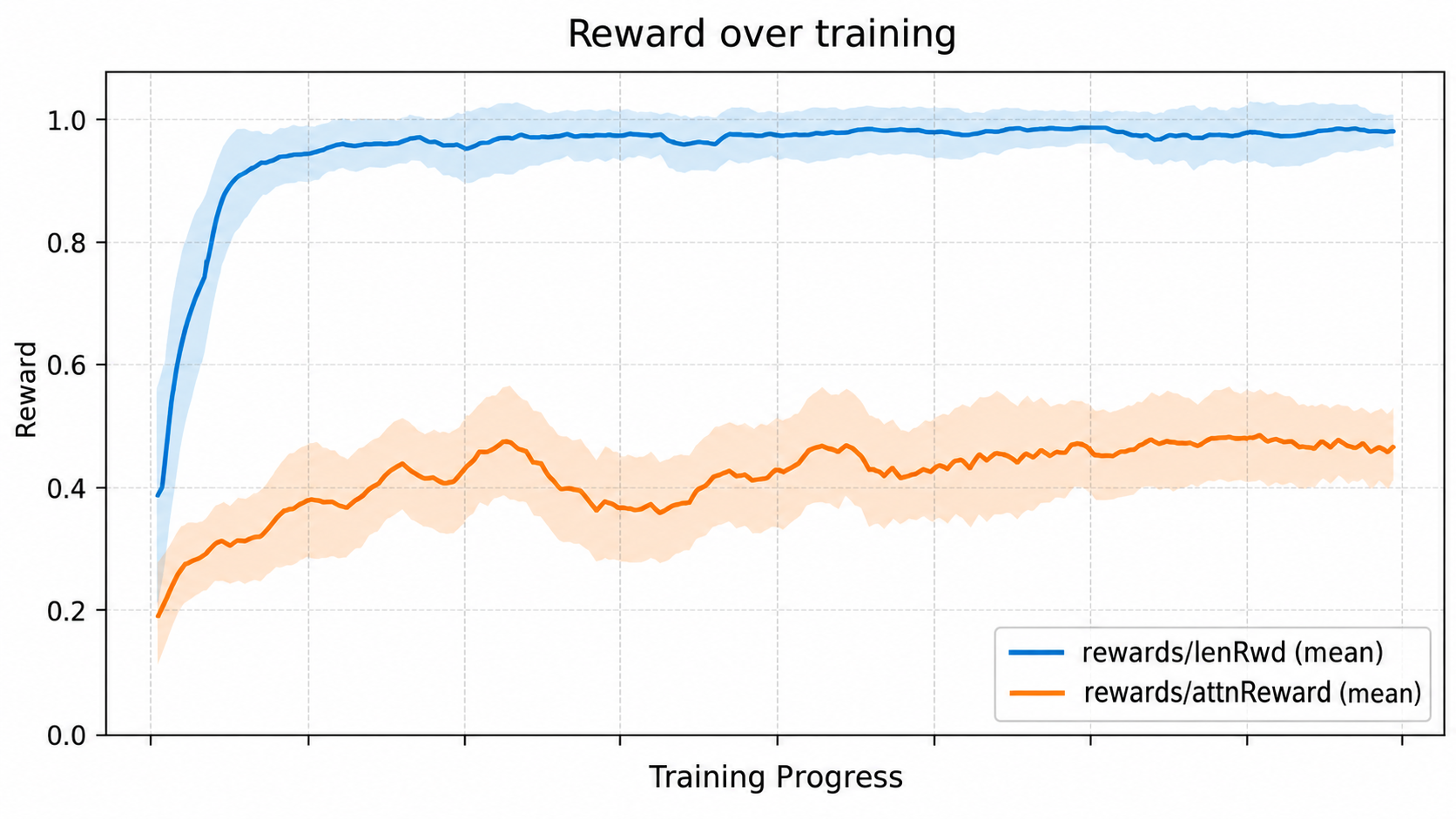}
    \caption{Progress of attention- and length-based rewards during GRPO training. The attention-based reward shows how the model progressively increases its attention toward both the cognitive component and counselor question segments, indicating improved focus and alignment over training steps. The length-based reward shows how the model learns to produce responses within the desired length range over training steps.}
    \vspace{-0.3in}
    \label{fig:attention_curve.png}
\end{figure}

\begin{figure}[h]
 \small   
\setlength{\abovecaptionskip}{0.1em} 
    \centering
    \includegraphics[width=3in]{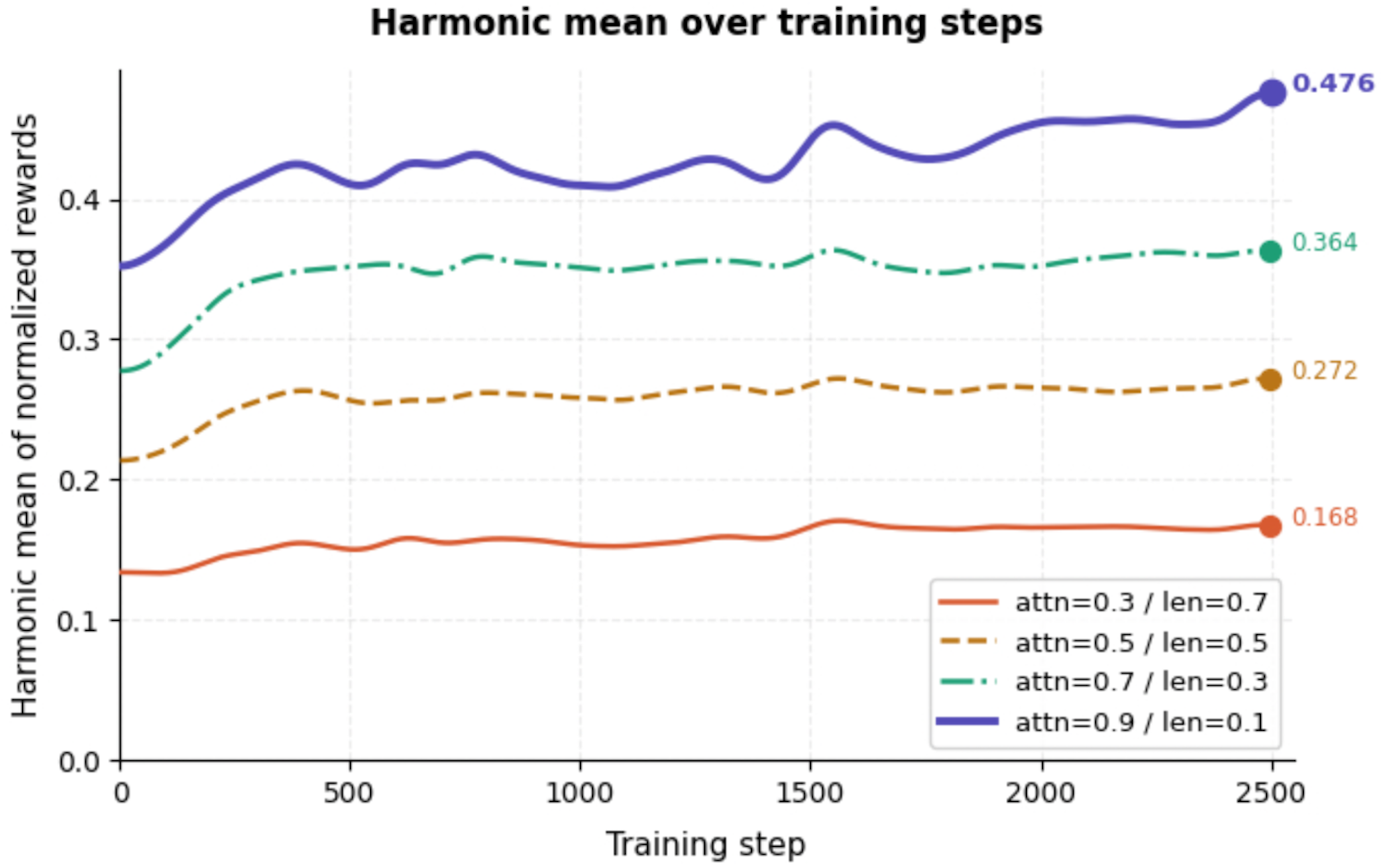}
    \caption{Harmonic mean of length and attention reward for different weight distributions. }
    \vspace{-0.2in}
    \label{fig:harmonic_mean_rewards}
\end{figure}

\subsubsection{CADG component importance: Role of SFT, DPO and GRPO: } Baseline model without any finetuning and relying only on prompting generates extraneous preambles along with the dialogues. Supervised instruction tuning (SFT) with GPT generated dialogues encouraged the model to mitigate this issue, as verified by human evaluation.
Direct Preference Optimization (DPO) further improves the model’s behavior on open-ended cognitive components by increasing the relative preference for targeted dialogues over non-targeted ones. As shown in Figure~\ref{fig:DPO_dial_reward}, the learned preference signal shifts toward the targeted open-ended cognition dialogues compared to other open-ended cognition dialogues.

Finally, the reward based optimizer (GRPO) enhances alignment of the dialogues on the targeted cognitive component, counselor question and the length constraint. This is evidenced in by the improved metrics in Table \ref{tab:updated_CADG_results} and corresponding reward curves in \ref{fig:attention_curve.png} 
, which show the model progressively optimizing toward higher rewards on these dimensions.

\subsection{GRPO reward ablation and weight tuning}
Our GRPO objective combines three reward signals: an attention reward that encourages the model to focus on both the cognitive component and counselor question segments in the prompt, and a length reward that penalizes deviations from the desired response length. We ablate these signals by training models with different subsets of rewards and report ROUGE, BLEU, and METEOR scores for cognitive-component and counselor-question alignment, as well as length compliance, in Table~\ref{tab:grpo_ablation}.Specifically, CConlyAttendRwd removes the counselor-question attention and length rewards, CQonlyAttendRwd retains only the counselor-question attention reward, and lengthOnlyRwd uses only the length reward, while CADG uses the full combination in weighted setting.
The ablation results show that each reward signal behaves as intended: CConlyAttendRwd achieves the highest ROUGE and METEOR scores with respect to the cognitive component text, CQonlyAttendRwd maximizes counselor-question ROUGE and METEOR, and lengthOnlyRwd (together with CADG) yields the strongest adherence to the target length constraint. 

However, qualitative inspection of CConlyAttendRwd and CQonlyAttendRwd responses reveals that many outputs closely mirror or directly copy the cognitive component or counselor-question prompts, respectively. The very high ROUGE scores in these settings thus indicate reward hacking: the model maximizes the attention reward by verbatim reuse of the prompt text rather than producing naturalistic,instruction aligned dialogue. In contrast, CADG—with a weighted combination of attention (cognitive component and counselor question) and length rewards—achieves lower cognitive component and counselor question ROUGE than CConlyAttendRwd and CQonlyAttendRwd, but generates instruction aligned and diverse dialogues that still satisfy cognitive, counselor, and length constraints, suggesting a better trade-off between alignment and naturalness.

Figure~\ref{fig:harmonic_mean_rewards} plots the harmonic mean of normalized attention and length rewards for different weight distributions, showing that higher attention weights (e.g., 0.9 attention / 0.1 length) achieve the best balance across both signals; this is the configuration used in our CADG setup. Figure~\ref{fig:attention_curve.png} 
further show that attention and length rewards steadily improve over training, confirming that GRPO can effectively shape the model’s focus and verbosity.

\subsection{LLM-as-judge quality evaluation}
\subsubsection{Human evaluation: }
We evaluated the quality of LLM judge response by annotating 10\% of the test set (51 instances ) with 5 Computer Science PhD student annotators. Each instances here are the human labeled cognitive component and the corresponding LLM-as-judge mapped sentence from the CADG generated dialogue. Annotators labeled the correctness of this mapping by labeling 0 if they are not aligned and  1 if aligned). Since the LLM-as-judge outputs were generally of high quality, the annotations were strongly skewed toward the positive class, with 47 out of 51 instances receiving a majority label of 1. Reviewers demonstrated high observed agreement, with an average agreement of 89.8\% with the majority label and unanimous agreement on 32 out of 51 instances.
Final labels were obtained via majority voting.

\begin{table}[t]
\caption{Confusion matrix comparing GPT-5 and GPT-5 Mini predictions across three classes (CADG worse, CADG better, CADG same). Diagonal entries (bold) indicate agreement. Agreement metrics show moderate categorical agreement (Cohen’s $\kappa$) and strong ranking consistency (Spearman $\rho$).}
\label{tab:gpt_agreement}
\centering
\footnotesize
\setlength{\tabcolsep}{1.8pt}
\renewcommand{\arraystretch}{0.95}
\begin{tabular}{lcccc}
\toprule
\textbf{GPT-5 $\backslash$ GPT-5 Mini} & \textbf{Worse } & \textbf{Better } & \textbf{Same } & \textbf{Row Total} \\
\midrule
\textbf{Worse }  & \textbf{20} & 2  & 5   & 27 \\
\textbf{Better } & 3  & \textbf{32} & 4   & 39 \\
\textbf{Same }   & 2  & 9  & \textbf{416} & 427 \\
\midrule
\textbf{Column Total} & 25 & 43 & 425 & 493 \\
\midrule
\multicolumn{5}{l}{\textbf{Cohen’s Kappa ($\kappa$): 0.5421 (Moderate agreement)}} \\
\multicolumn{5}{l}{\textbf{Spearman Correlation ($\rho$): 0.5528 (Moderate, $p < 1e{-40}$)}} \\
\bottomrule
\vspace{-0.2in}
\end{tabular}

\end{table}

\subsubsection{LLM-judge sensitivity}
We assess the sensitivity of our LLM-as-judge setup by comparing GPT-5 Mini against GPT-5 as reference. Table~\ref{tab:gpt_agreement} shows a confusion matrix over three outcome classes—\emph{CADG worse}, \emph{CADG better}, and \emph{CADG same}—where rows correspond to GPT-5 predictions and columns to GPT-5 Mini predictions. The diagonal entries are relatively large, indicating frequent agreement between the two judges across all three classes. We further report Cohen’s $\kappa$ and Spearman’s $\rho$, which show moderate categorical agreement and consistent system rankings, respectively. These results suggest that GPT-5 Mini provides judgments that are broadly consistent with GPT-5 while being substantially cheaper to run, making it a practical choice for large-scale evaluation.

\section{Conclusion}

In this work, we present a novel framework for SUD AI patient dialogue
generation in LLMs. In this framework, the generated dialogue reflects patient's
underlying cognitive states, which are derived from patient's history. The
overall framework consists of two modules: cognitive component detector, and
cognitive aligned dialogue generator, the first module extracts seventeen
different cognitive components related to SUD patient behavior from the patient
history, and then the second module generate patient dialogues that are aligned
with these extracted cognitive components. To develop these modules we use
smaller LLM models, by combining knowledge distillation from high-capacity
teacher model through supervised fine-tuning, preference optimization from
human-annotations, and attention-guided reward optimization. Extensive
evaluations using automatic scores like BERTScore, ROUGE, METEOR and BLEU, and
LLM-as-judge hit-metrics validates the effectiveness of the proposed framework;
specifically, that cognitively informed fine-tuning and reward optimization 
substantially improves cognitive realization and alignment over a generic 
instruction-tuned baseline.

\section{Limitations and Ethical Considerations}

We use publicly available Reddit posts on substance use and recovery, which may contain sensitive mental health information. Posts were collected via Reddit’s API using PRAW. We removed PII, usernames, and identifiers, rewrote posts as standardized third-person synthetic narratives, and discarded the originals. As per Reddit policy no raw data or derivatives are released. These synthetic patient histories are intended only for controlled research on cognitive alignment in dialogue generation. As such, the system is not intended to replace professional judgment, provide medical advice, or be used in therapeutic settings.

\section{Acknowledgment}
We acknowledge the contribution from Dr. Ellen and her group at Indiana University for providing domain knowledge in SUD counseling for us to pursue this work.

\bibliographystyle{IEEEtran}
\bibliography{custom}

\end{document}